\documentclass{article}

\usepackage{microtype}
\usepackage{hanging}
\usepackage{graphicx}
\usepackage{subcaption}
\usepackage{booktabs}
\usepackage{hyperref} 

\PassOptionsToPackage{hyphens}{url}\usepackage{hyperref}

\usepackage{algorithm}
\usepackage{algorithmic}
\usepackage{tikz}
\usepackage{pgfplots}
\usepackage[]{mdframed}

\usepackage[accepted]{icml2025}

\usepackage{amsmath}
\usepackage{amssymb}
\usepackage{mathtools}
\usepackage{amsthm}

\usepackage[capitalize,noabbrev]{cleveref}

\theoremstyle{plain}

\theoremstyle{definition}

\theoremstyle{remark}

\usepackage[textsize=tiny]{todonotes}

\begin{document}

\twocolumn[
\icmltitle{Pheromone-based Learning of Optimal Reasoning Paths}



\icmlsetsymbol{equal}{*}

\begin{icmlauthorlist}
\icmlauthor{Anirudh Chari}{comp,sch1,equal}
\icmlauthor{Aditya Tiwari}{comp,equal}
\icmlauthor{Richard Lian}{comp,sch1}
\icmlauthor{Suraj Reddy}{comp,sch1}
\icmlauthor{Brian Zhou}{comp,sch2}
\end{icmlauthorlist}

\icmlaffiliation{comp}{a37.ai, San Francisco, CA, USA}
\icmlaffiliation{sch1}{Massachusetts Institute of Technology, Cambridge, MA, USA}
\icmlaffiliation{sch2}{Harvard University, Cambridge, MA, USA}

\icmlcorrespondingauthor{Anirudh Chari}{anichari@mit.edu}

\icmlkeywords{Machine Learning, ICML}

\vskip 0.3in
]



\printAffiliationsAndNotice{\icmlEqualContribution} 

\begin{abstract}
Large Language Models (LLMs) have demonstrated remarkable reasoning capabilities through chain-of-thought prompting, yet discovering effective reasoning methods for complex problems remains challenging due to the vast space of possible intermediate steps. We introduce Ant Colony Optimization-guided Tree of Thought (ACO-ToT), a novel algorithm that combines ACO with LLMs to discover optimal reasoning paths for complex problems efficiently. Drawing inspiration from Hebbian learning in neurological systems, our method employs a collection of distinctly fine-tuned LLM ``ants'' to traverse and lay pheromone trails through a centralized tree of thought, with each ant's movement governed by a weighted combination of existing pheromone trails and its own specialized expertise. The algorithm evaluates complete reasoning paths using a mixture-of-experts-based scoring function, with pheromones reinforcing productive reasoning paths across iterations. Experiments on three challenging reasoning tasks (GSM8K, ARC-Challenge, and MATH) demonstrate that ACO-ToT performs significantly better than existing chain-of-thought optimization approaches, suggesting that incorporating biologically inspired collective search mechanisms into LLM inference can substantially enhance reasoning capabilities.

\end{abstract}

\section{Introduction}

\usetikzlibrary{shapes,arrows,positioning,calc,backgrounds,fit,decorations.pathreplacing}

\begin{figure*}[!ht]
    \centering
    \begin{subfigure}[b]{0.12\linewidth}
        \resizebox{0.9\linewidth}{!}{
            \begin{tikzpicture} [
                node distance=0.75cm,
                input/.style = {draw, ellipse},
                output/.style = {draw, ellipse, fill=green!40},
                thought/.style = {draw, rounded corners, minimum width=1.25cm, minimum height=0.75cm},
            ]
                \node[input, minimum width=2cm, minimum height=1cm] (input) {
                    \scalebox{1.3}{Input}
                };
            
                \node[thought, below= of input] (t1) {};
                \node[thought, below= of t1] (t2) {};
                \node[thought, below= of t2] (t3) {};
            
                \node[output, below= of t3, minimum width=2.6cm, minimum height=1.3cm] (output) {
                    \scalebox{1.3}{Output}
                };

                \draw[->] (input) -- (t1);
                \draw[->] (t1) -- (t2);
                \draw[->] (t2) -- (t3);
            \end{tikzpicture}
        }
        \caption{Chain of Thoughts (CoT)}
    \end{subfigure}
    \hfill
    \begin{subfigure}[b]{0.35\textwidth}
        \resizebox{\linewidth}{!}{
        \begin{tikzpicture} [
            node distance=0.75cm,
            input/.style = {draw, ellipse},
            output/.style = {draw, ellipse, fill=green!40},
            thought/.style = {draw, rounded corners, minimum width=1.25cm, minimum height=0.75cm},
            lgthought/.style = {draw, rounded corners, fill=green!10, minimum width=1.25cm, minimum height=0.75cm},
            gthought/.style = {draw, rounded corners, fill=green!40, minimum width=1.25cm, minimum height=0.75cm},
            rthought/.style = {draw, rounded corners, fill=red!10, minimum width=1.25cm, minimum height=0.75cm},
        ]
            \scalebox{1.3}{
                \node[input, minimum width=2cm, minimum height=1cm] (input) {
                    Input
                };
            }
        
            \node[gthought, below= of input] (s1) {};
            \node[thought, right= of s1] (t1) {};
            \node[thought, left= of s1] (t2) {};
            
            \node[thought, below= of s1] (s2) {};
            \node[gthought, right= of s2] (m1) {};
            \node[thought, left= of s2] (m2) {};
            \node[gthought, left= of m2] (m3) {};
            \node[thought, right= of m1] (m4) {};
            
            \node[thought, below=of s2] (s3) {};
            \node[gthought, right= of s3] (d1) {};
            \node[thought, right= of d1] (d3) {};
            \node[thought, left=2.75cm of s3] (d2) {};
        
            \node[below=0.5 of s3] (ellipses) {
                .....
            };
        
            \draw[->] (input) -- (s1);
            \draw[->] (input) -- (t1);
            \draw[->] (input) -- (t2);
            \draw[->] (t2) -- (m3);
            \draw[->] (s1) -- (m1);
            \draw[->] (s1) -- (s2);
            \draw[->] (s1) -- (m2);
            \draw[->] (t1) -- (m4);
            \draw[->] (m1) -- (s3);
            \draw[->] (m1) -- (d1);
            \draw[->] (m1) -- (d3);
            \draw[->] (m3) -- (d2);
        
            \node[output, below= of ellipses, minimum width=2.6cm, minimum height=1.3cm] (output) {   
                \scalebox{1.3}{Output}
            };
        
            \draw[->] (ellipses) -- (output);
        
        \end{tikzpicture}
        }
        \caption{Tree of Thoughts (ToT)}
    \end{subfigure}
    \hfill
    \vrule
    \hfill
    \begin{subfigure}[b]{0.35\textwidth}
        \resizebox{\linewidth}{!}{
        \begin{tikzpicture} [
            node distance=0.75cm,
            input/.style = {draw, ellipse},
            output/.style = {draw, ellipse, fill=green!40},
            thought/.style = {draw, rounded corners, minimum width=1.25cm, minimum height=0.75cm},
            lgthought/.style = {draw, rounded corners, fill=green!10, minimum width=1.25cm, minimum height=0.75cm},
            gthought/.style = {draw, rounded corners, fill=green!40, minimum width=1.25cm, minimum height=0.75cm},
            rthought/.style = {draw, rounded corners, fill=red!10, minimum width=1.25cm, minimum height=0.75cm},
            ant/.style = {draw, circle, fill=yellow!20, scale=0.6, line width=0.2}
        ]
            \scalebox{1.3}{
                \node[input, minimum width=2cm, minimum height=1cm] (input) {
                    Input
                };
            }
        
            \node[lgthought, below= of input] (s1) {};
            \node[thought, right= of s1] (t1) {};
            \node[gthought, left= of s1] (t2) {};
            
            \node[lgthought, below= of s1] (s2) {};
            \node[lgthought, right= of s2] (m1) {};
            \node[rthought, left= of s2] (m2) {};
            \node[lgthought, left= of m2] (m3) {};
            \node[rthought, right= of m1] (m4) {};
            
            \node[thought, below=of s2] (s3) {};
            \node[gthought, right= of s3] (d1) {};
            \node[lgthought, right= of d1] (d3) {};
            \node[rthought, left=2.75cm of s3] (d2) {};
        
            \node[below=0.5 of s3] (ellipses) {
                .....
            };
        
            \draw[->, line width=0.33mm] (input) -- (s1);
            \draw[->, line width=0.33mm] (input) -- (t1);
            \draw[->, line width=0.33mm] (input) -- (t2);
            \draw[->, line width=0.25mm] (t2) -- (m3) node[midway, ant] {$a_0$};
            \draw[->, line width=0.35mm] (s1) -- (m1);
            \draw[->, line width=0.20mm] (s1) -- (s2);
            \draw[->, line width=0.15mm] (s1) -- (m2);
            \draw[->, line width=0.15mm] (t1) -- (m4);
            \draw[->, line width=0.2mm] (m1) -- (s3);
            \draw[->, line width=0.5mm] (m1) -- (d1) node[midway, ant] {$a_1$};
            \draw[->, line width=0.2mm] (m1) -- (d3);
            \draw[->, line width=0.1mm] (m3) -- (d2);

            \node[output, below= of ellipses, minimum width=2.6cm, minimum height=1.3cm] (output) {   
                \scalebox{1.3}{Output}
            };
        
            \draw[->] (ellipses) -- (output);
        
        \end{tikzpicture}
        }
        \caption{Weighted "Pheromone" ToT} \label{fig:acototoverview}
    \end{subfigure}
    
    \caption{Comparison of approaches to complex reasoning problems with LLMs. Each rectangular node represents \textit{a thought}---an intermediate reasoning step to solve a larger problem. On the right, our method utilizes ``ants'' traversing between connected reasoning steps (depicted as circles) to strengthen productive reasoning steps across iterations via pheromone trails. See algorithmic implementation for \ref{fig:acototoverview} in Figure \ref{fig:algflowchart}.} \label{fig:banner}
\end{figure*}
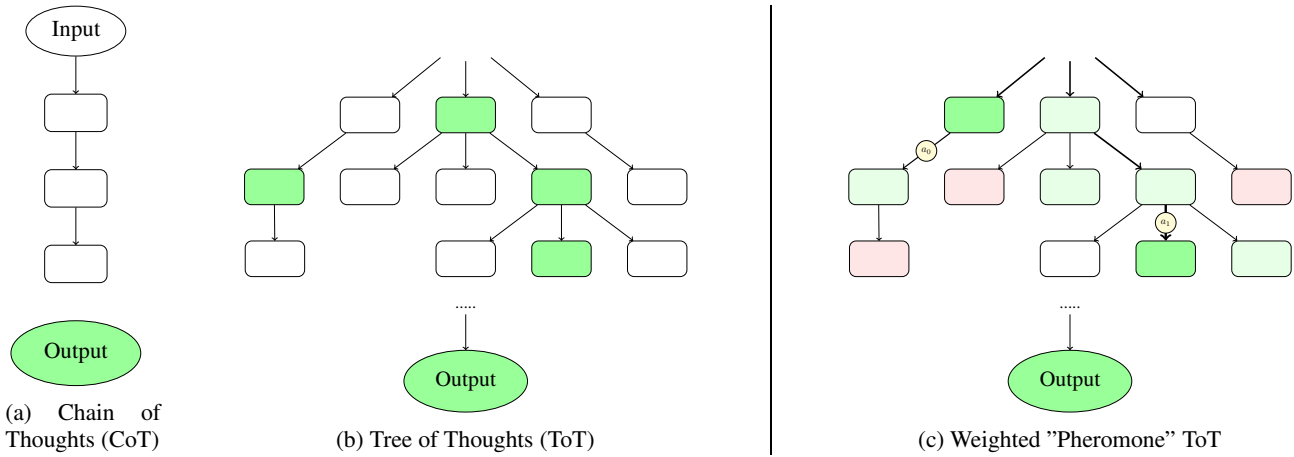

Large language models (LLMs) have demonstrated remarkable capabilities in emulating human-like behavior across diverse tasks \citep{brown_language_2020}. These models process sequences of tokens through attention mechanisms, achieving strong performance on mathematical, logical, and commonsense reasoning benchmarks \citep{wei_chain--thought_2023}. However, their widespread adoption faces two critical limitations: computational efficiency and reasoning accuracy \citep{chowdhery_palm_2022}. Specifically, their reasoning capabilities remain constrained by token-level processing within the attention architecture, necessitating enhanced reasoning mechanisms for applications requiring exploration, strategic planning, or deterministic initial steps \cite{yao_tree_2023}.

Recent research has shifted from black-box approaches toward interpretable methodologies grounded in cognitive science principles \cite{ling_program_2017}. Chain-of-Thought (CoT) prompting exemplifies this transition by explicitly generating intermediate reasoning steps before producing final outputs \cite{wei_chain--thought_2023}. This method implements a verification mechanism where the model evaluates its own reasoning, significantly improving accuracy on complex tasks \cite{kojima_large_2023}.
However, naive CoT effectively only considers a single approach to solving a problem. Meanwhile, humans may consider many possible approaches to a problem before deciding upon a productive reasoning method \citep{newell_human_1972}, which points toward the notion of CoT ``optimization.''

To build upon CoT's initial success and address this key limitation, researchers have developed various specialized frameworks for structured exploration of the \textit{reasoning space}. Notable derivatives include Tree-of-Thoughts (ToT) \cite{yao_tree_2023}, Graph-of-Thoughts (GoT) \cite{besta_graph_2024}, and Iterative Reasoning Preference Optimization (IRPO) \cite{bai_constitutional_2022}. These methodologies transcend CoT's linear progression by implementing branching, backtracking, and systematic space exploration, more accurately reflecting human problem-solving strategies. However, as reasoning tasks grow in complexity, these methods incur substantial computational overhead due to the exponential growth in intermediate reasoning steps required for accurate solutions \cite{yao_tree_2023}.

Hebbian learning \cite{Hebb1949} provides a paradigm for understanding pathway optimization in biological neural networks. The principle, commonly expressed as ``neurons that fire together, wire together,'' describes how repeated activation of sequences of neurons strengthens synaptic connections between those neurons, establishing preferential pathways for efficient signal propagation \cite{lowel_selection_1992}.

We propose adapting this neurological optimization principle to LLM reasoning through Ant Colony Optimization (ACO), which we claim sufficiently resembles Hebbian learning mechanics. Our implementation, ACO-ToT, employs specialized fine-tuned LLMs as artificial ants traversing a reasoning space. These LLM-ants deposit virtual pheromones proportional to reasoning quality. The system implements a probabilistic path selection mechanism, where pheromone concentration influences path choice while maintaining exploration-exploitation balance through stochastic selection. This collective intelligence approach gradually converges on optimal reasoning strategies while pruning inefficient paths \cite{blum_ant_2005}.

Hence, our main contributions are: 
\begin{itemize}
    \item A neuroscience-inspired paradigm for search and optimization across the natural-language reasoning space via pheromone mechanics,
    \item \textbf{ACO-ToT}, a mixture-of-experts algorithm applying the above paradigm for dynamic CoT optimization using artificial LLM-based ants, and
    \item Extensive experimental validation of ACO-ToT demonstrating a mean absolute accuracy improvement of \textbf{16.6\%} over existing approaches.
\end{itemize}
Section \ref{background} contextualizes the integration of Hebbian learning principles with Tree of Thoughts prompting. Sections \ref{methods} and \ref{theoretical} detail our implementation methodology and analyze theoretical properties, including computational complexity. Section \ref{experiments} presents our experimental protocol using the GSM8K, ARC-Challenge, and MATH datasets, while Section \ref{results} provides comprehensive results and analysis for comparison with other flagship methods and ablation studies for accuracy optimization. Section \ref{related} describes related works in CoT prompting, LLM inference optimization, biological inspiration in AI, mixture of experts, and prompting techniques for increasing reasoning. Finally, Section \ref{conclusion} concludes our paper with a summary of findings and suggests future directions.

\section{Background}\label{background}






\subsection{Chain of Thought}
Chain-of-thought (CoT) prompting enables language models to break down complex reasoning tasks into intermediate steps, significantly improving their problem-solving capabilities \citep{wei_chain--thought_2023}. Given an input $x$ and desired output $y$, CoT introduces a sequence of intermediate thoughts $z_1,...,z_n$ that bridge the reasoning gap. Each thought $z_i$ represents a coherent language sequence sampled from the distribution $z_i \sim \pi_\theta(z_i|x,z_1,...,z_{i-1})$, where $\pi_\theta$ denotes the language model with parameters $\theta$.
The key innovation of CoT lies in its ability to decompose multi-step problems into manageable intermediate steps, allowing models to allocate computational resources according to problem complexity. This decomposition provides interpretable insights into the model's reasoning process and enables debugging of incorrect solutions. Empirically, CoT prompting has demonstrated significant improvements across arithmetic, commonsense, and symbolic reasoning tasks \citep{DBLP:journals/corr/abs-2201-07207}.
\subsection{Tree of Thought}
Tree of Thought (ToT) extends CoT by enabling exploration of multiple reasoning paths simultaneously \citep{yao_tree_2023}. Rather than generating a single chain, ToT maintains a tree where each node represents a state $s=[x,z_{1:i}]$ containing the input and sequence of thoughts thus far. At each state, ToT generates $k$ candidate next thoughts and evaluates their promise toward solving the problem.
The ToT framework comprises four key components: (1) thought decomposition into appropriate semantic units, (2) thought generation through sampling or sequential proposal, (3) state evaluation via independent scoring or comparative voting, and (4) tree search algorithms like breadth-first or depth-first search to explore the reasoning space. This deliberate exploration allows ToT to overcome limitations of left-to-right decoding by considering multiple paths and backtracking when necessary.
\subsection{Ant Colony Optimization}
Ant Colony Optimization (ACO) is a metaheuristic inspired by the foraging behavior of ant colonies \citep{782657}. In ACO, artificial ants traverse a graph representing possible solutions, depositing pheromone trails proportional to solution quality. The probability $p_{ij}$ of an ant choosing edge $(i,j)$ is given by:
\begin{equation}
p_{ij} = \frac{(\tau_{ij})^\alpha(\eta_{ij})^\beta}{\sum_{l\in N_i}(\tau_{il})^\alpha(\eta_{il})^\beta}
\end{equation}
where $\tau_{ij}$ is the pheromone level, $\eta_{ij}$ is a heuristic value, $\alpha$ and $\beta$ are parameters controlling their relative importance, and $N_i$ is the set of available next nodes. After each iteration, pheromone levels are updated according to:
\begin{equation}
\tau_{ij} \leftarrow (1-\rho)\tau_{ij} + \sum_{k=1}^m \Delta\tau_{ij}^k
\end{equation}
where $\rho$ is the evaporation rate and $\Delta\tau_{ij}^k$ is the pheromone deposited by ant $k$.
The pheromone reinforcement mechanism in ACO resembles Hebbian learning in biological neural networks, where repeatedly activated synaptic pathways become stronger over time \citep{ye_deepaco_2023}. This parallel suggests that pheromone-based path selection could naturally extend to reasoning-space search in a ToT, where promising cognitive trajectories are strengthened through repeated traversal while unproductive paths decay through evaporation. Figure \ref{fig:banner} visually highlights the main differences between CoT, ToT, and ACO-ToT during reasoning-space exploration.

\subsection{Mixture of Experts}
Mixture of Experts (MoE) for LLMs leverages multiple specialized models to collaboratively work on complex tasks. In the context of LLMs, MoE architectures consist of a set of ``expert" models, each fine-tuned for specific reasoning types or domains, and a routing mechanism that determines how to weigh expert outputs for a given input \citep{si_getting_2023}. The advantages of incorporating MoE in LLM reasoning motivate us to consider LLMs as ``ants'' during reasoning-space exploration \citep{yao_tree_2023}.


\section{Methods}\label{methods}

\subsection{Algorithm Overview}

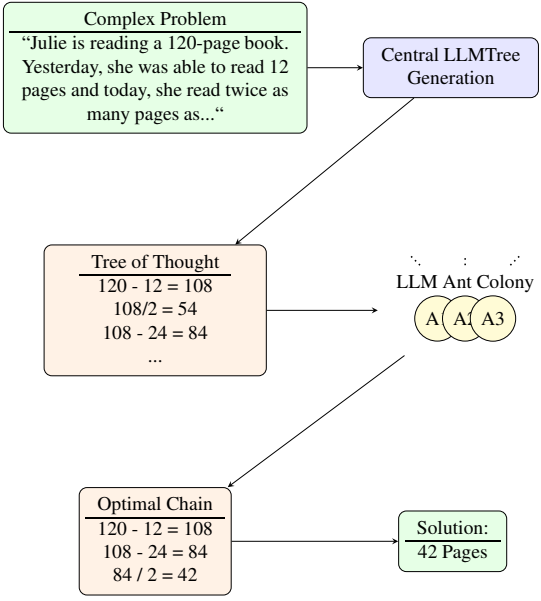
\begin{figure}
\resizebox{0.9\linewidth}{!}{
\begin{tikzpicture}[
    node distance=2cm,
    box/.style={draw, rounded corners, minimum width=2.5cm, minimum height=1.2cm},
    process/.style={draw, rounded corners, fill=blue!10},
    data/.style={draw, rounded corners, fill=green!10},
    ant/.style={draw, circle, fill=yellow!20},
    thought/.style={draw, rounded corners, fill=orange!10},
    >=stealth
]
    \node[data] (input) {
        \begin{tabular}{c} 
        Complex Problem \\
        \hline
        ``Julie is reading a 120-page book. \\ Yesterday, she was able to read 12 \\ pages and today, she read twice as \\ many pages as...``\\
        \end{tabular}
        };
    
    \node[process, right=1cm of input] (llm) {
        \begin{tabular}{c} Central LLMTree \\ Generation \end{tabular}
    };
    
    \node[thought, below=of input, minimum width=4cm] (tot) {
        \begin{tabular}{c}
            Tree of Thought\\
            \hline
            120 - 12 = 108\\
            108/2 = 54\\
            108 - 24 = 84\\
            ...
        \end{tabular}
    };
    
    \node[right=2cm of tot] (ants) {
        \begin{tabular}{c}
            LLM Ant Colony\\
            \begin{tikzpicture}[scale=0.5]
                \node[ant] at (0,0) {A1};
                \node[ant] at (1,0) {A2};
                \node[ant] at (2,0) {A3};
            \end{tikzpicture}
        \end{tabular}
    };
    
    \node[thought, below=of tot] (path) {
        \begin{tabular}{c}
            Optimal Chain\\
            \hline
            120 - 12 = 108\\
            108 - 24 = 84\\
            84 / 2 = 42
        \end{tabular}
    };
    
    \node[data, right=3cm of path] (solution) {\begin{tabular}{c} Solution: \\ \hline 42 Pages \end{tabular}};
    
    \draw[->] (input) -- (llm);
    \draw[->] (llm) -- (tot);
    \draw[->] (tot) -- (ants);
    \draw[->] (ants) -- (path);
    \draw[->] (path) -- (solution);
    
    \draw[dotted, thick] (ants) -- ++(1,1);
    \draw[dotted, thick] (ants) -- ++(0,1);
    \draw[dotted, thick] (ants) -- ++(-1,1);
\end{tikzpicture}
}
\caption{Example procedure for a math problem from GSM8K. The central LLM is prompted for an initial ToT, which is next explored by fine-tuned ant LLMs to discover an optimal reasoning path, and then computed for a final result. See Algorithm \ref{alg:mainaco} for general procedure.} \label{fig:algflowchart}
\end{figure}

We propose ACO-guided Tree of Thought (ACO-ToT), a novel algorithm that combines the exploration capabilities of ToT with the collective intelligence of ACO to discover optimal reasoning paths. The algorithm employs a collection of specialized LLM ``ants'' to traverse a reasoning graph generated by a central LLM, converging on high-quality CoTs through iterative exploration and reinforcement. For a general overview of the algorithm, see Figure \ref{fig:algflowchart}.

\subsection{Reasoning Graph Construction}

Given a problem input $x$, we first prompt a central LLM $\pi_c$ to generate a tree of thought $\mathcal{T}$. We augment $\mathcal{T}$ with special start node $s_0$ and end node $s_f$ to form a directed graph $G=(V,E)$, where vertices $V$ represent reasoning states and edges $E$ represent transitions between states. Each state $s_i \in V$ contains the problem input and accumulated reasoning chain: $s_i = [x, z_{1:i}]$. Psuedocode for ToT generation is found as Algorithm \ref{alg:tot}.

\subsection{LLM Ant Colony}

Motivated by MoE architectures, we maintain a collection of $m$ distinctly fine-tuned LLMs $\{\pi_1,...,\pi_m\}$ that serve as specialized ``ants''. Each LLM $\pi_k$ is fine-tuned on different aspects of reasoning, providing diverse expertise. At each timestep $t$, ant $k$ at state $i$ chooses its next state $j$ with probability:

\begin{equation}
    p_{ij}^k = \frac{(\tau_{ij})^\alpha(h_{ij}^k)^\beta}{\sum_{l\in N_i}(\tau_{il})^\alpha(h_{il}^k)^\beta}
\end{equation}

where $\tau_{ij}$ is the pheromone level on edge $(i,j)$, $h_{ij}^k$ is the heuristic value computed by prompting LLM $\pi_k$ for its assessment of state $j$, and $N_i$ is the set of available next states.

\subsection{Path Evaluation and Pheromone Update}

For a complete path $P=(s_0,...,s_f)$, i.e., a chain of thought, we compute its quality $Q(P)$ as:

\begin{equation}
Q(P) = w_1C(P) + w_2L(P) + w_3M(P)
\end{equation}

where:
\begin{itemize}
\item $C(P)$ is the semantic coherence measured via embedding cosine similarity between consecutive states
\item $L(P)$ is a length penalty term: $-\log(|P|)$
\item $M(P)$ is a mixture-of-experts score: $\frac{1}{m}\sum_{k=1}^m \pi_k(P)$
\item $w_1,w_2,w_3$ are weights
\end{itemize}
This quality function roughly captures the logic, complexity, and agreeability of a given reasoning path.

Pheromone levels are updated according to:

\[
\tau_{ij} \leftarrow (1-\rho)\tau_{ij} + \sum_{k=1}^m \Delta\tau_{ij}^k
\]

where $\Delta\tau_{ij}^k = Q(P_k)$ if edge $(i,j)$ is in ant $k$'s path $P_k$, and 0 otherwise. Across iterations, these updates yield further exploitation of high-quality reasoning strategies.

\begin{algorithm}[htb]
\caption{ACO-guided Tree of Thought}
\begin{algorithmic}[1]
\REQUIRE Problem $x$, central LLM $\pi_c$, ant LLMs $\{\pi_1,\ldots,\pi_m\}$, iterations $T$
\ENSURE Optimal chain of thought $z^*$

\STATE \textbf{Initialize reasoning graph:}
\STATE $G = (V,E) \leftarrow \text{GenerateToT}(\pi_c, x)$
\STATE $\tau_{ij} \leftarrow \tau_0$ for all $(i,j) \in E$

\STATE \textbf{Main ACO loop:}
\FOR{$t = 1$ to $T$}
    \STATE \textbf{Initialize ant paths:}
    \STATE $P_k \leftarrow [s_0]$ for $k = 1,\ldots,m$
    
    \STATE \textbf{Construct solutions:}
    \WHILE{$\exists k: s_f \notin P_k$}
        \FORALL{ant $k$ with incomplete path}
            \STATE $i \leftarrow \text{last state in } P_k$
            \STATE $j \leftarrow \text{SampleNextState}(\pi_k, \tau, h)$ \COMMENT{Using Eq. 3}
            \STATE $P_k \leftarrow P_k \cup \{j\}$
        \ENDFOR
    \ENDWHILE
    
    \STATE \textbf{Evaluate paths and update pheromones:}
    \FORALL{edge $(i,j) \in E$}
        \STATE $\tau_{ij} \leftarrow (1-\rho)\tau_{ij}$
        \FOR{$k = 1$ to $m$}
            \IF{$(i,j) \in P_k$}
                \STATE $\tau_{ij} \leftarrow \tau_{ij} + Q(P_k)$ \COMMENT{Using Eq. 4}
            \ENDIF
        \ENDFOR
    \ENDFOR
\ENDFOR

\STATE \textbf{return} $\text{ExtractBestPath}(G, \tau)$
\end{algorithmic}
\label{alg:mainaco}
\end{algorithm}

\subsection{Convergence and Path Extraction}

The algorithm iterates until either reaching a maximum iteration count $T$ or satisfying a convergence criterion based on path diversity across iterations. The final optimal chain of thought $z^*$ is extracted from the path with highest pheromone levels in the graph. This path can then be used by the central LLM $\pi_c$ to generate the final solution.


\section{Theoretical Properties}\label{theoretical}

\subsection{Classical Ant Colony Optimization}
Ant Colony Optimization (ACO) is a metaheuristic inspired by ant foraging behavior, widely applied to combinatorial optimization problems. Here we analyze its key theoretical properties informing our application to reasoning-space application.

\textbf{Convergence}
ACO converges to optimal solutions under specific conditions. In general, with pheromone update rules and evaporation rates appropriately chosen according to the given problem, the algorithm asymptotically approaches the global optimum while avoiding local optima. Formally, this convergence can be expressed as:

\begin{equation}
\lim_{i \to \infty} P(z^*, i, k) = 1
\end{equation}

where $P$ represents the probability of ant $k$ finding optimal path $z^*$ on iteration $i > i^*$, with $i^*$ being the iteration where the first optimal solution is discovered \cite{dorigo_aco_book_2004}. In practice, convergence can be induced by allowing a max-iteration time instead of waiting for a completely stable state, as ACO-ToT implements here. Techniques like elitism, which prioritize top solutions in pheromone updates, can accelerate this convergence. See \citet{dorigo_aco_book_2004} for more rigorous analysis.

\textbf{Exploration vs. Exploitation}
The balance between exploration and exploitation in ACO is governed by pheromone dynamics. The parameters $\alpha$ and $\beta$ control this balance, with $\alpha$ weighting the influence of pheromone trails (exploitation) and $\beta$ weighting heuristic information (exploration). The evaporation rate further modulates this balance: higher rates promote exploration while lower rates reinforce exploitation. This mechanism is particularly crucial in complex solution spaces with multiple local optima. For more rigorous analysis, see \cite{dorigo_aco_2006}.

\textbf{Computational Complexity}
ACO exhibits polynomial complexity in both the number of ants and problem size. For instance, in the traveling salesman problem, solution construction per ant scales as $O(n^2)$ for $n$ cities, with pheromone updates requiring equivalent complexity, working out a weighted optimal path, which can found to be similar to finding an optimal path through reasoning considering MoE weights. While efficient for moderate-sized problems, the algorithm's iterative nature can become computationally intensive for large instances, though notably, this can be significantly mitigated through parallelization.

\subsection{ACO-ToT Analysis}
The integration of ACO with Tree of Thoughts introduces unique computational considerations that must be carefully balanced against performance gains.

\textbf{API Costs}
The primary cost metric in LLM-based implementations is API call volume. For ACO-ToT, each ant requires $N$ intermediate thoughts per explored path over $t$ iterations, resulting in $A \cdot N \cdot t$ total LLM calls for $A$ ants. This compounds with the base ToT overhead of $\sum_{i=1}^{d} n^i$ calls for tree generation, where $d$ is tree depth and $n$ is the branching factor.

\textbf{Computational Overhead}
Parallel LLM execution introduces significant resource demands compared to standard ToT. Each ant requires dedicated GPU memory for model loading, while concurrent inference can strain computational resources. To manage these costs, we implement:

\begin{enumerate}
    \item A maximum iteration count $T$ to force early convergence
    \item Efficient parallelization through batched LLM calls
    \item Multi-GPU distribution for memory optimization
\end{enumerate}

These constraints align well with typical ToT implementations, where reasoning paths can be bounded around the optimal 5-6 steps \cite{yao_tree_2023}, enabling ACO-ToT to operate efficiently within the smaller ToT. We further study the practical capabilities within ablation trials found in future sections.

\section{Experiments}\label{experiments}

We wish to empirically prove that an implementation of ACO-ToT challenges standard and flagship models of reasoning in LLMs, through testing on multiple databases. The following section details information about the setup used, datasets and baselines tested against, and additional metrics stored for ablation studies in the Results section.

\subsection{Experimental Setup}
For all experiments, we use Llama-70b as our base language model, with five distinctly fine-tuned LLM experts serving as ants:
\begin{itemize}
    \item Mathematical reasoning expert (fine-tuned on ProofNet)
    \item Scientific reasoning expert (fine-tuned on ScienceQA)
    \item Logical deduction expert (fine-tuned on LogiQA)
    \item Common sense reasoning expert (fine-tuned on CSQA)
    \item Domain-specific expert (fine-tuned on task-specific data)
\end{itemize}

Implementation details:
\begin{itemize}
    \item Number of LLM ants $m = 5$
    \item Pheromone evaporation rate $\rho = 0.1$
    \item Exploitation vs exploration weights $\alpha = 1, \beta = 2$
    \item Path quality weights $w_1 = 0.4$ (coherence), $w_2 = 0.3$ (length), $w_3 = 0.3$ (expert consensus)
    \item Maximum iterations $T = 10$ or until convergence
    \item Convergence threshold: path stability for 3 consecutive iterations
\end{itemize}
All experiments were run using 8 NVIDIA A100 GPUs with 80GB memory.

\subsection{Datasets}
\textbf{GSM8K}
Grade school math word problems containing 7.5K training and 1K test examples. Each problem requires multi-step reasoning to arrive at a numerical answer \cite{cobbe2021gsm8k}. We use the standard train/test split and evaluate using exact match accuracy.

\textbf{ARC-Challenge}
Science questions consisting of 2,590 training and 1,172 test examples \cite{clark2018thinksolvedquestionanswering}. Each question is multiple choice with 4 options. We evaluate using accuracy on the challenge set.

\textbf{MATH}
Competition math problems across different categories, with 7,500 training and 5,000 test problems \cite{hendrycks2021measuringmathematicalproblemsolving}. Problems require advanced mathematical reasoning and formal notation.

\subsection{Baselines}

We compare ACO-ToT against three strong baselines representing different approaches to LLM reasoning:

\textbf{Chain of Thought (CoT)}, as introduced previously, generates intermediate reasoning steps sequentially to bridge input and output. We use the standard CoT prompting approach with Llama-70b as described in Wei et al. (2023).

\textbf{Tree of Thought (ToT)}, as introduced previously, explores multiple reasoning paths by maintaining a tree of intermediate thoughts. We implement the BFS variant of ToT using the same thought decomposition and evaluation strategies as described in Yao et al. (2023).

\textbf{Iterative Reasoning Preference Optimization (IRPO)} represents the current state-of-the-art learning-based approach. IRPO iteratively optimizes preference between competing CoT candidates by training on winning vs. losing reasoning steps using a modified DPO loss with an additional negative log-likelihood term. While other learning-based methods require extensive training procedures, IRPO achieves strong performance through efficient preference optimization over multiple iterations. We use the authors' implementation with their reported best hyperparameters.

All baselines use the same Llama-70b base model for fair comparison. For IRPO and ToT, we use the same maximum iteration count $(T=10)$ and convergence criteria as ACO-ToT.

\subsection{Evaluation Metrics}
For each task, we measure the success rate of the model, by percentage of problems solved correctly. To monitor the algorithm we measure both convergence speed, or number of iterations until convergence, and path quality metrics, given by average path length, mean coherance score, and expert agreement rate.

\section{Results and Analysis}\label{results}

\subsection{Main Results}
Table 1 presents the performance comparison across all tasks:

\begin{table}[!htb]
\centering
\caption{Performance comparison across tasks (accuracy \%)}
\begin{tabular}{l|ccc}
\hline
Method & GSM8K & ARC-Challenge & MATH \\
\hline
CoT & 55.6 & 77.8 & 12.5 \\
ToT & 68.3 & 82.1 & 16.4 \\
IRPO & 81.6 & 86.7 & 20.8 \\
ACO-ToT (Ours) & \textbf{84.2} & \textbf{88.9} & \textbf{22.6} \\
\hline
\end{tabular}
\end{table}

ACO-ToT demonstrates substantial performance gains across all evaluation tasks, achiving absolute improvements of 28.6\%, 11.1\%, and 10.1\% over standard CoT prompting on GSM8K, ARC-Challenge, and MATH respectively. These improvements are particularly noteworthy when compared to reinforcement learning approaches like IRPO. IRPO shows significant improvements in early iterations, with gains of 17.5\%, 4.9\%, 3.1\%, and 0.5\% across its first four iterations before performance saturates. Similarly, ACO-ToT converges to high-quality solutions within 6-8 iterations for standard problems, extending to 10-12 iterations for more complex MATH problems, and yielding consistently higher quality solutions than IRPO.

This rapid convergence manifests in a characteristic performance curve: steep improvement in the first 3-4 iterations followed by asymptotic stabilization, indicating efficient exploration of the reasoning space. The algorithm's performance-to-computation ratio is particularly favorable, as it achieves state-of-the-art results without the computational overhead of RL training procedures or the extensive sampling requirements of other methods.

\begin{figure}[t]
\begin{tikzpicture}
\begin{axis}[
    width=0.48\textwidth,
    height=0.35\textwidth,
    xlabel={Number of Iterations},
    ylabel={Performance Metrics},
    xmin=0, xmax=10,
    ymin=0, ymax=100,
    legend pos=south east,
    grid=major,
    grid style={gray!30},
    legend style={font=\small},
    legend cell align=left
]

\addplot[thick,blue] coordinates {
    (0,55.6) (1,73.1) (2,78.0) (3,81.1) (4,81.6) (5,81.8) (6,81.9) (7,82.0) (8,82.0) (9,82.0)
};

\addplot[thick,red] coordinates {
    (0,65) (1,75) (2,82) (3,85) (4,86) (5,86) (6,86) (7,86) (8,86) (9,86)
};

\addplot[thick,green] coordinates {
    (0,70) (1,78) (2,81) (3,82) (4,82) (5,82) (6,82) (7,82) (8,82) (9,82)
};

\legend{Accuracy \%, Expert Agreement \%, Coherence Score \%}
\end{axis}
\end{tikzpicture}
\hfill
\begin{tikzpicture}
\begin{axis}[
    width=0.48\textwidth,
    height=0.35\textwidth,
    xlabel={Number of Iterations},
    ylabel={Path Properties},
    xmin=0, xmax=10,
    ymin=0, ymax=10,
    legend pos=north east,
    grid=major,
    grid style={gray!30},
    legend style={font=\small},
    legend cell align=left
]

\addplot[thick,red] coordinates {
    (0,6.2) (1,5.4) (2,4.8) (3,4.6) (4,4.5) (5,4.4) (6,4.4) (7,4.4) (8,4.4) (9,4.4)
};

\addplot[thick,orange] coordinates {
    (0,3.1) (1,2.8) (2,2.6) (3,2.5) (4,2.4) (5,2.4) (6,2.4) (7,2.4) (8,2.4) (9,2.4)
};

\legend{Average Path Length, Pheromone Concentration Ratio}
\end{axis}
\end{tikzpicture}
\caption{Convergence analysis of ACO-ToT showing (left) performance metrics and (right) path properties across iterations. The algorithm typically converges after 3-4 iterations, with accuracy improving from 55.6\% to 81.6\%, expert agreement reaching 86\%, and coherence scores stabilizing at 82\%. Path lengths decrease and stabilize at 4.4 steps on average, while the ratio of pheromone concentration between optimal and suboptimal paths reaches 2.4.}
\label{fig:convergence}
\end{figure}
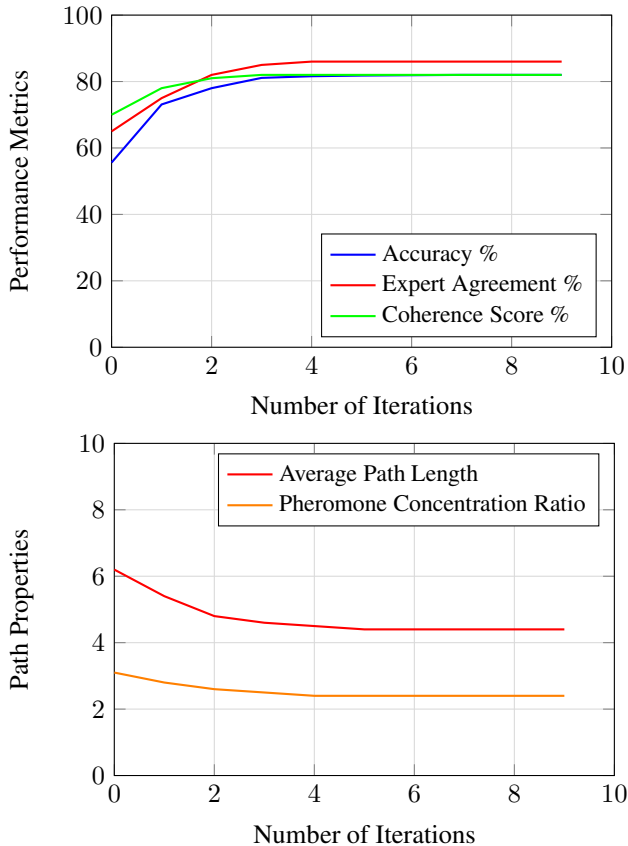

\subsection{Analysis of Path Properties}
We analyze the properties of converged reasoning paths based on previous evaluation metrics.

Path length distribution allows us to consider the supposed complexity of problems given by datasets: on average, GSM8K had 4.8 steps ($\sigma = 1.3$), ARC-Challenge had 4.2 steps ($\sigma = 1.1$), and MATH had 6.1 steps ($\sigma = 1.6$). Additionally, our 82\% average agreement rate between experts on optimal paths also resulted in higher agreement. In later data analysis, we found that agreement rate correlates strongly with solution accuracy with an $r$-value of $0.78$. Additional metrics included pheromone concentration, which showed that optimal paths show 2.8$\times$ higher pheromone concentration vs suboptimal. The concentration gradient steepened over iterations and, when corroborated by previous results, demonstrates that the iterative approach to refining CoTs was largely effective.

\subsection{Ablation Studies}
We conduct extensive ablation studies to analyze the impact of different components in ACO-ToT, eventually settling on the following hyperparameters as an optimal trade-off between mixture-of-experts idea diversity and computational cost without sacrificing the accuracy of final answers.

\textbf{Number of LLM Ants}
Performance saturates around 5 experts, suggesting this is a happy medium between MoE diversity and computational cost (Table \ref{abel:experts}).

\textbf{Pheromone Parameters}
The optimal balance between exploitation ($\alpha=1.0$) and exploration ($\beta=2.0$) yields best performance across tasks (Table \ref{abel:alpvbet}). This result follows from expecting large trees and allowing differing opinions to attempt to come to better convergences, e.g., Science Expert would be encouraged to explore the tree instead of following a Mathematical Expert and lay pheromone elsewhere, motivating a search for a global minimum instead of possible local minima.

\textbf{Scoring Components}
All three scoring components contribute to performance, with coherence and MoE being particularly important (Table \ref{abel:scoring}).

\textbf{Expert Diversity}
Diverse expert specialization (mathematical, scientific, logical, common sense, and domain-specific) outperforms homogeneous expert configurations (Table \ref{abel:diversity}). This suggests that a heavier MoE approach was rewarded with more generally acceptable reasoning steps towards a solution.

\subsection{Summary of Findings}
Our results demonstrate several key findings, mainly: ACO-ToT consistently outperforms existing methods across all tasks, with particularly strong gains on complex reasoning problems like GSM8K and MATH. The algorithm typically converges within 6-8 iterations, with more complex problems requiring more iterations. However, if required, performance improves rapidly in early iterations before plateauing. 

Our ablation trials found that expert diversity is crucial -- having specialized LLMs for different aspects of reasoning leads to better exploration of the solution space. The pheromone mechanism laid by these experts effectively guides the search, with optimal parameters $\alpha = 1.0$ and $\beta = 2.0$ suggesting a focus on exploration allows the ant LLMs to converge timely while exploring reasoning through the tree. As for optimization, we found that all three components of the scoring function (coherence, length penalty, and mixture of experts) contribute meaningfully to performance.

\section{Related Work}\label{related}
\subsection{Chain-of-Thought and Tree-of-Thought Prompting}
Recent advances in prompting strategies have revealed emergent reasoning in large language models (LLMs). Chain-of-Thought (CoT) prompting, introduced by \citep{wei_chain--thought_2023}, demonstrates that explicitly generating intermediate reasoning steps (e.g.,``Let's think step by step'') improves performance on arithmetic, commonsense, and symbolic reasoning tasks. However, CoT suffers from three key limitations: (1) its linear, step-by-step structure propagates errors in long reasoning chains, (2) it lacks mechanisms to backtrack or explore alternative paths, and (3) manual curation of high-quality exemplars is labor-intensive.
To address error propagation, Self-Consistency \citep{wang_voyager_2023} aggregates multiple CoT paths via majority voting, improving accuracy on GSM8K. However, this method scales poorly due to its brute-force sampling strategy. Concurrently, Zero-Shot CoT \citep{kojima_large_2023} prompted models to "think step by step", but it retained CoT's linearity and struggled with tasks requiring global planning (e.g., The Game of 24, which achieved a low success rate).
The Tree-of-Thought (ToT) framework \citep{yao_tree_2023} introduces non-linear reasoning by exploring multiple paths as a search tree, enabling backtracking and look-ahead. ToT improves Game of 24 success rates to 74\% with GPT-4, but its fixed tree structure limits scalability to complex tasks with exponential search spaces.

\subsection{Optimization Techniques for Language Model Inference}
Optimizing LLM inference for reasoning tasks has focused on decoding strategies and feedback mechanisms. IRPO \citep{bai_constitutional_2022} iteratively refines reasoning paths using human preference data, improving MATH benchmarks. However, IRPO's use of human feedback limits its real-world applicability. Guided Decoding \citep{li_guiding_nodate} incorporates domain-specific heuristics (e.g., mathematical rules) during generation but requires task-specific tuning and failed to generalize.
Automatic CoT \citep{zhang_automatic_2022} automates exemplar generation using LLMs to produce diverse reasoning chains, matching manual CoT performance on 10 reasoning tasks. However, error correction remains challenging due to noisy self-generated chains. Additionally, Active Prompting \citep{diao_active_2024} uses uncertainty-aware exemplar selection to improve CoT reliability but requires labeled validation data. ACS \citep{arias_adaptive_2024} introduces an adaptive contrastive search to balance diversity and coherence in text generation. However, its focus on creativity limits its utility for structured reasoning.
\subsection{Biologically-Inspired Optimization for AI}
Biologically inspired algorithms like ACO have been adapted for neural architectures but rarely for reasoning. DeepACO \citep{ye_deepaco_2023} combines ACO with deep RL for combinatorial optimization, outperforming traditional solvers on TSP. However, it focuses on low-level graph traversal rather than semantic reasoning. Next, the Neural Architecture Search with ACO \citep{lankford_neural_2024} optimizes model designs but ignores in-context reasoning dynamics. In the Cumulative Reasoning paper \citep{zhang_cumulative_2024}, researchers employ iterative hypothesis refinement similar to ant foraging, improving factuality in open-domain question and answer. However, it lacks explicit exploration-exploitation mechanisms for path optimization.


\subsection{Reasoning Enhancement Techniques}
Recent work has targeted specific reasoning failures. Faithful CoT \citep{lyu_faithful_2023} grounds reasoning steps in external knowledge bases, reducing hallucinations. STaR \citep{zelikman_star_2022} bootstraps reasoning via self-training but requires fine-tuning. Least-to-Most Prompting \citep{zhou_least--most_2023} decomposes complex problems into subquestions, improving compositional generalization but struggling with interdependent steps.

\section{Conclusion}\label{conclusion}
In this work we presented ACO-guided Tree of Thought (ACO-ToT), a novel approach to enhancing language models' reasoning capabilities by combining ant colony optimization with chain-of-thought prompting. Our method demonstrates that incorporating neuroscience-inspired collective search mechanisms into language model inference can substantially improve problem-solving performance. The algorithm achieves significant improvements over existing approaches across various reasoning benchmarks, with accuracy gains of 28.6\% on GSM8K, 11.1\% on ARC-Challenge, and 10.1\% on MATH compared to standard chain-of-thought prompting.

While effective, our approach has several limitations. First, the computational cost of running multiple LLM experts and iterations may be prohibitive for some applications. Second, the performance depends heavily on the quality of expert diversity and pheromone parameter tuning. Third, the current implementation requires manual specification of scoring functions and convergence criteria.

Future work could explore several promising directions, including the automated tuning of pheromone parameters and scoring weights, integration with other search algorithms like MCTS or A*, development of more sophisticated expert specialization strategies, investigation of meta-learning approaches to improve convergence speed, and an extension to multi-modal reasoning tasks.

\section*{Impact Statement}
This work introduces a novel approach to optimizing language model reasoning paths through biologically-inspired collective search mechanisms. By adapting ant colony optimization principles to guide chain-of-thought reasoning, our method enables more robust and efficient problem-solving across complex mathematical, scientific, and logical reasoning tasks. While the immediate implications are primarily academic, focused on improving automated reasoning systems' accuracy and efficiency, potential downstream applications could include educational support systems, automated theorem proving, and scientific discovery assistance. However, these applications would require additional safeguards and careful consideration of fairness, bias, and transparency before deployment in real-world settings. As with any advancement in AI reasoning capabilities, we acknowledge the broader societal implications while maintaining our current focus on fundamental research in controlled environments, emphasizing the importance of responsible development practices for any future real-world applications.


\bibliography{example_paper}
\bibliographystyle{icml2025}

\newpage
\onecolumn

\appendix
\section{Implementation Details}
\subsection{Hyperparameter Settings}
\begin{itemize}
    \item Number of LLM ants $m = 5$
    \item Pheromone evaporation rate $\rho = 0.1$
    \item Exploitation vs exploration weights $\alpha = 1, \beta = 2$
    \item Path quality weights $w_1 = 0.4, w_2 = 0.3, w_3 = 0.3$
    \item Maximum iterations $T = 10$ or until convergence
    \item Convergence threshold: path stability for 3 consecutive iterations
\end{itemize}

\subsection{Expert Models}
The five distinctly fine-tuned LLM experts:
\begin{itemize}
    \item Mathematical reasoning expert (fine-tuned on ProofNet)
    \item Scientific reasoning expert (fine-tuned on ScienceQA)
    \item Logical deduction expert (fine-tuned on LogiQA)
    \item Common sense reasoning expert (fine-tuned on CSQA)
    \item Domain-specific expert (fine-tuned on task-specific data)
\end{itemize}

\subsection{Computational Resources}
All experiments were run using 8 NVIDIA A100 GPUs with 80GB memory. Average runtime per task:
\begin{itemize}
    \item GSM8K: 8.2s
    \item ARC-Challenge: 6.5s  
    \item MATH: 12.4s
\end{itemize}

\subsection{Prompt Information}

The prompt used for all three datasets was modeled after the GSM8K dataset, and is as follows:

\ttfamily

\hspace{1em} \begin{mdframed}
Imagine you are trying to solve a math problem with a step-by-step approach. At each step, you should propose a single next step to solve the problem involving a single arithmetic option. If there are multiple options for how to proceed, you should generate up to 3 options.

The format of the problem is as below, follow this format only

\vspace{1em}

Input: XXXX

Steps taken so far: YYYY

Output: ZZZZ

\vspace{1em}

NOTE: The options should not be sequential or connected with each other, each option should be in a way that it can be evaluated independently. Don't jump to the result directly.

IMPORTANT: **MAKE SURE NOT TO HAVE THE DIRECT ANSWER IN YOUR POSSIBLE STEPS OUTPUT, JUST MAKE ONE STEP AT A TIME.**

\vspace{1em}

Solved Example:

Example 1

Input: "Jasper will serve charcuterie at his dinner party. He buys 2 pounds of cheddar cheese for \$10, a pound of cream cheese that cost half the price of the cheddar cheese, and a pack of cold cuts that cost twice the price of the cheddar cheese. How much does he spend on the ingredients?"

Steps take so far: [Calculate the price of cheddar cheese which is \$10 (given)]

\vspace{1em}

Output: Possible independent steps:

1) Calculate the price of cold cuts which is 2*10 = \$20.

2) Calculate the price of cream cheese which is 10/2 = \$5 per pound.

\vspace{1em}

Example 2

Input: "Weng earns \$12 an hour for babysitting. Yesterday, she just did 50 minutes of babysitting. How much did she earn?"

Steps taken so far: [None]

\vspace{1em}

Output: Possible next steps:

1) Convert the minutes of babysitting to hours.

2) Convert the wage per hour to wage per minute.

\vspace{1em}

Example 3

Input: "James writes a 3-page letter to 2 different friends twice a week. How many pages does he write a year?"

Steps taken so far: [Number of letter written to 1 friend in a week = 2 as he writes twice a week]

\vspace{1em}

Output: Possible next steps:

1) Number of letter written to 2 friends in a week = 2*2 = 4 letters a week.

2) Calculate the number of pages written to 1 friend in a week = 2*3 = 6 pages.

\vspace{1em}

Now give the possible independent next steps for the below question, making one specifically numerical step at a time to solve the problem, without jumping to a proposed answer solution or repeating previous answer steps.

\vspace{1em}

Input: "[problem here]"

Steps taken so far: [previous steps here]

\vspace{1em}

Output:

1)

\end{mdframed}

\rmfamily

\subsection{Tree-of-Thought Generation Algorithm}


\begin{algorithm}[htb]
    \begin{algorithmic}[1]
    \REQUIRE Problem $x$, central LLM $\pi_c$, thought generator $G()$ max depth $D$, branches $B$
    \ENSURE Tree-of-Thoughts $\mathcal{T}$

    \STATE \textbf{Initialize graph and storage:}
    \STATE $\mathcal{T} = (V, E), V \leftarrow p$ \COMMENT{Tree Root}
    \STATE $u \leftarrow [\{p\}, \{\},...,\{\}$ \COMMENT{Set to manage unique thoughts}

    \STATE \textbf{Main generation loop:}
    \FOR{$d = 0$ to $D - 1$}
        \FORALL{node $\nu$ at depth $d$}
            \STATE $p \leftarrow \text{FindAncestors}(\nu, \mathcal{T})$
            \STATE $o \leftarrow G(\pi_c, x, p)$
            \STATE $t \leftarrow \text{ExtractThoughts}(o, B)$
            \FORALL{thought $\tau$ in $t$}
                \IF{$\tau$ not in $u_d$}
                    \STATE $u_d \leftarrow u_d \cup \{\tau\}$
                    \STATE $V \leftarrow t, E \leftarrow (p_d, t))$ \COMMENT{Adds thought at $d+1$}
                \ENDIF
            \ENDFOR
        \ENDFOR
    \ENDFOR
    \STATE \textbf{return} $\mathcal{T}$
    \end{algorithmic}
    \caption{Algorithm for generation Tree of Thoughts $\mathcal{T}$ based on problem $x$.}\label{alg:tot}
\end{algorithm}

\section{Ablation Results}

Tables contained here include information from the ablation tests in order to pick optimal hyper-parameters for highest accuracy. Information here is analyzed above:

\begin{table}[h]
\centering
\begin{subtable}{0.5\linewidth}
\caption{Effect of number of LLM experts on performance (\%)}
\centering
\begin{tabular}{lccc}
\hline
\# Experts & GSM8K & ARC & MATH \\
\hline
2 & 75.2 & 82.4 & 16.1 \\
3 & 77.8 & 84.1 & 17.9 \\
5 & 81.6 & 86.7 & 20.8 \\
7 & 81.9 & 86.9 & 20.9 \\
8 & 82.0 & 87.0 & 21.0 \\
\hline
\end{tabular}
\label{abel:experts}
\end{subtable}\hfill
\begin{subtable}{0.5\linewidth}
\caption{Impact of exploitation ($\alpha$) vs exploration ($\beta$) weights}
\centering
\begin{tabular}{ccccc}
\hline
$\alpha$ & $\beta$ & GSM8K & ARC & MATH \\
\hline
0.5 & 2.0 & 77.3 & 83.2 & 18.4 \\
1.0 & 2.0 & 81.6 & 86.7 & 20.8 \\
2.0 & 2.0 & 79.1 & 84.5 & 19.2 \\
1.0 & 1.0 & 76.8 & 82.9 & 17.9 \\
\hline
\end{tabular}
\label{abel:alpvbet}
\end{subtable}
\end{table}

\begin{table}[!htb]
\centering
\begin{subtable}{0.4\linewidth}
\caption{Ablation of scoring function components}
\centering
\begin{tabular}{lccc}
\hline
Components & GSM8K & ARC & MATH \\
\hline
Full (C+L+M) & 81.6 & 86.7 & 20.8 \\
Only Coherence (C) & 75.3 & 81.2 & 16.4 \\
Only Length (L) & 72.1 & 79.8 & 15.2 \\
Only MoE (M) & 76.8 & 82.5 & 17.3 \\
C+L & 77.9 & 83.4 & 18.1 \\
C+M & 79.2 & 84.9 & 19.4 \\
L+M & 76.1 & 81.8 & 16.9 \\
\hline
\end{tabular}
\label{abel:scoring}
\end{subtable}\hfill
\begin{subtable}{0.4\linewidth}
\centering
\caption{Impact of expert specialization}
\begin{tabular}{lccc}
\hline
Expert Configuration & GSM8K & ARC & MATH \\
\hline
Full Diversity & 81.6 & 86.7 & 20.8 \\
Math-only & 79.2 & 82.1 & 19.4 \\
Science-only & 76.8 & 84.3 & 17.2 \\
Logic-only & 77.5 & 83.1 & 18.1 \\
Random Mix & 75.9 & 81.4 & 16.8 \\
\hline
\end{tabular}
\label{abel:diversity}\vspace{1em}
\end{subtable}\hspace{5em}

\end{table}

\end{document}